\documentclass[conference]{IEEEtran}
\IEEEoverridecommandlockouts
\usepackage{amsmath,amssymb,amsfonts}
\usepackage{algorithmic}
\usepackage{graphicx}
\usepackage{textcomp}
\usepackage{xcolor}
\usepackage{booktabs}
\usepackage{multirow}
\usepackage{bigdelim}
\def\BibTeX{{\rm B\kern-.05em{\sc i\kern-.025em b}\kern-.08em
		T\kern-.1667em\lower.7ex\hbox{E}\kern-.125emX}}

\begin{document}
	\newcommand{\todos}[1]{\textcolor{red}{\textbf{S}: #1}}
\newcommand{\todok}[1]{\textcolor{green}{\textbf{K}: #1}}
\newcommand{\todol}[1]{\textcolor{blue}{\textbf{L}: #1}}
	\title{	Assessment of Neural Networks for Stream-Water-Temperature Prediction	}
	
        \author{\IEEEauthorblockN{Stefanie Mohr}
 		\IEEEauthorblockA{\textit{Chair of Theoretical Computer Science} \\
 			\textit{Technical University of Munich}\\
 			Munich, Germany \\
 			0000-0002-8630-3218}
 		\and
        \IEEEauthorblockN{Konstantina Drainas}
 		\IEEEauthorblockA{\textit{Chair of Aquatic Systems Biology} \\
 			\textit{Technical University of Munich}\\
 			Munich, Germany \\
 			0000-0001-6771-2123}

 		\and
 		\IEEEauthorblockN{Prof. Dr. J\"urgen Geist}
 		\IEEEauthorblockA{\textit{Chair of Aquatic Systems Biology} \\
 			\textit{Technical University of Munich}\\
 			Munich, Germany \\
 			0000-0001-7698-3443}
 	}
	
	\maketitle
	
	\begin{abstract}
	    Climate change results in altered air and water temperatures. 
	    Increases affect physicochemical properties, such as oxygen concentration, and can shift species distribution and survival, with consequences for ecosystem functioning and services. 
		These ecosystem services have integral value for humankind and are forecasted to 
		alter under climate warming. 
		A mechanistic understanding of the drivers and magnitude of expected changes is essential in identifying system resilience and mitigation measures. 
		In this work, we present a selection of state-of-the-art Neural Networks (NN) for the prediction of water temperatures in six streams in Germany. 
		We show that the use of methods that compare observed and predicted values, exemplified with the Root Mean Square Error (RMSE), is not sufficient for their assessment. Hence we introduce additional analysis methods for our models to complement the state-of-the-art metrics. 
		These analyses evaluate the NN's robustness, possible maximal and minimal values, and the impact of single input parameters on the output.
		We thus contribute to understanding the processes within the NN and help applicants choose architectures and input parameters for reliable water temperature prediction models. \\

	\end{abstract}
	
	\begin{IEEEkeywords}
		neural network, prediction, water temperature, climate change, verification, evaluation, robustness
	\end{IEEEkeywords}

	\section{Introduction}
Recent years have already shown the impact of climate change on various organisms, among them keystone species from aquatic ecosystems, such as macroinvertebrates. 
These animals, which by convention can be retained by a 500 $\mu$m mesh net and lack a backbone, are often overlooked in the public discourse, even though they are abundant in nearly all types of streams and rivers all over the world \cite{Hauer.2017b} and have great relevance for the functioning of stream ecosystems \cite{Wallace.1996}. 
They play a very important role in energy flow and hence also for higher levels of the food chain \cite{Wallace.1996}. 
However, certain macroinvertebrate species and many fish species have stringent cold-water temperature requirements, which makes them particularly vulnerable to warmer water conditions. 
Air temperature has a strong influence on the temperature of rivers, particularly in broad and flat headwater streams that show tight coupling to atmospheric processes due to the high ratio of stream surface to water depth \cite{Gomi02}. 
Consequently, increases in air temperature will also affect water temperature, leading to reduced cold water patches as possible habitats \cite{kuhn2021assessing}. 
In Germany, an increase by 1.5\textdegree C in the annual mean temperature between 1881 and 2018 has already been detected \cite{Umweltbundesamt.November2019}.
Similarly, an increase in mountain-lake water temperature between 0.1\textdegree C  and  1.1\textdegree C per decade has been observed \cite{kuefner2020evaluating}. 
For ecological stakeholders, the knowledge of whether and how much temperature will change is highly important for the introduction of preventive policies, e.g. investing in shady riparian vegetation to reduce stream water temperature \cite{trimmel2018can}. 

To be able to plan and execute such preventive policies, good and reliable models for water temperature prediction are essential. 
Their creation has two main advantages: 
Firstly, it will help to predict future changes in temperature and thus helps in choosing reasonable prevention and mitigation methods.
Secondly, based on a few years of measurements, water temperature in streams can be predicted instead of measured. 
Even though measurements are more precise, predictions would not only reduce costs and effort but also establish possibilities for researchers as well as for ecological stakeholders all over the world to work with data that up to now have been hard to obtain.

Different approaches for water temperature prediction exist and have already been used for various aquatic systems. 
Linear regression, for example, is often used to describe the relationship between air and water temperature \cite{ Caissie06, Harvey.2011, Krider.2013, Pilgrim.1998, Rabi15, Smith81}.
However, when plotting air against water temperature, physical effects concerning high and low temperatures lead to a non-linear, s-shaped relationship between the two parameters \cite{Mohseni.1999}.
Additionally, atmospheric conditions, topography, stream discharge, bedform, and riparian vegetation play a relevant role as an influence on water temperature \cite{Caissie06, Beschta97}.
Hence, linear regression based on air temperature can not be considered optimal for water temperature prediction. 
Addressing the non-linear relationship, Neural Networks (NN) seem very promising for reliable water temperature predictions. 

We use the knowledge gained in hitherto studies to create state-of-the-art NN for water temperature prediction and introduce analysis methods besides the commonly used metrics. Since these metrics exclusively compare observed and predicted values, we do not rely on them but complement them by our additional analysis methods, to assess and hence choose proper NN for water temperature prediction. 

As much as NN are known to be effective in learning, they are also known to be vulnerable to perturbations \cite{papernot2016limitations,akhtar2018threat}.
In our setting, a small change in the output is not necessarily a problem. 
However, users of river temperature models, such as hydrologists or employees in water management offices, may not be aware that there is a difference between a NN-model and a classical model, such as one using linear regression.
That is why we need a more thorough analysis of the NN to evaluate its behavior more critically.
For this application, we introduce three methods: { \textit{Robustness-},} \textit{Min/Max-} and \textit{Impact-Analysis}.

The \textit{Robustness-Analysis} mimics the approach that is already known for classification \cite{DBLP:conf/atva/ChengNR17,DBLP:conf/atva/Ehlers17}. 
It evaluates the impact of a small perturbation on the output value.
In a linear regression model, a perturbation of the input would be linearly transformed to the output.
For NN, this is not as simple because their behavior is not linear. 
That is why the calculation of a robustness-value is a problem that has gained a lot of interest in the last few years \cite{papernot2016limitations,akhtar2018threat,DBLP:conf/atva/ChengNR17,DBLP:conf/atva/Ehlers17}. 
\cite{singh2019abstract} that evaluates a NN in terms of its robustness.

Furthermore, the \textit{Min/Max-Analysis} is meant to determine minimal and maximal values that the NN can take. 
For classical models, this value is almost obvious or at least can be evaluated by simple calculus.
NN make this more difficult due to their complicated structure.
However, we adopt the idea of \cite{szegedy2014intriguing} that generates specific input vectors to the NN that fulfill any pre-defined property. 
Even though the application was different, we can use a slightly adapted approach here. 

The purpose of the \textit{Impact-Analysis} is to help future developers of prediction models. 
The target groups who will want to use our approach may not be computer scientists themselves, but hydrologists or biologists.
Thus, we need a method that visualizes clearly which input values were most important for the decision-making of the NN.
On the one hand, this can help in adapting the model, on the other hand, it may also result in insights into the correlation between stream temperatures and other features.
Therefore, we use a similar approach to \cite{zurada1994sensitivity} to determine which input features contribute most to calculating the output value.

We aim to introduce assessment methods for non-experts in the field of NN. Therefore, we not only implement methods with understandable output but also use easily accessible data, to not limit our approach to specific measurement requirements. 
Our data were collected at different streams in Germany, which were selected as described in Section~\ref{data}. 
Currently, we have an individual model for each stream trained on data from one measurement site each. 
This means that the input data was not gathered along the whole length of the streams but at one certain measurement site per stream.

We summarize our contributions as follows:
\begin{itemize}
	\item We create state-of-the-art NN for six different streams in Germany 
	\item We show that metrics like the RMSE are not sufficient to assess NN for water temperature prediction properly
	\item We introduce thorough analysis methods for regression problems in NN 
	\begin{itemize}
	    \item to compare it to classical approaches
	    \item to be able to choose robust architectures and input combinations for reliable predictions
	    \item to make it more understandable to the user
	\end{itemize} 
\end{itemize} 
	\section{State of the art} \label{sota}
\subsection{Water temperature prediction} \label{sota_wtp}
As changes in water temperature have been identified as a key component of aquatic ecosystem health, their accurate prediction becomes increasingly important. 
Hence, several approaches for water temperature prediction already exist. 
The simplest approach is linear regression, presuming a linear relationship between air and water temperature \cite{Caissie06, Krider.2013, Pilgrim.1998, Rabi15, Smith81}. 
Another approach is stochastic modeling, e.g., multiple regression analysis \cite{Caissie06, Caissie.1998, AhmadiNedushan07}, second-order Markov processes \cite{Caissie.1998}, Box and Jenkins time-series models \cite{Caissie.1998, AhmadiNedushan07}, and second-order autoregressive models \cite{AhmadiNedushan07}. 
Machine learning approaches, including Gaussian process regression, decision trees, and NN, have been tested, too \cite{Zhu.2019}.
In comparison, NN have shown either well \cite{Zhu.2019} or even best \cite{Rabi15, Chenard.2008b} performance and are becoming increasingly popular \cite{Rabi15, Zhu.2019, Chenard.2008b, Piotrowski.2015, HadzimaNyarko.2014}.

The studies using NN for water temperature prediction assessed their models based on bias, Coefficient of Determination \cite{Chenard.2008b}, Coefficient of Correlation, Coefficient of Efficiency, Adjusted Coefficient of Efficiency \cite{Rabi15}, Mean Absolute Error, Willmott Index of Agreement \cite{Zhu.2019}, Mean Square Error \cite{Piotrowski.2015}, and the Root Mean Square Error (RMSE) \cite{Rabi15, Zhu.2019, Chenard.2008b}. Even though there is a wide variety in these metrics, they all mainly compare observed and predicted values, not considering the underlying processes and the reliability of the NN's predictions.\\ 

Also, these metrics do not consider input parameters and their impact on the output, besides changes in prediction accuracy.

Nonetheless, several approaches have tried to improve water temperature prediction by additional input parameters that are supposed to complement air temperature from current and previous days as input parameters: runoff/relative change in flow \cite{AhmadiNedushan07, Zhu.2019}, which is the part of the effective precipitation that flows into the stream \cite{.20210622T10:23:51.000Z}, declination of sun \cite{Piotrowski.2015}, soil temperature \cite{StHilaire2000}, riparian vegetation \cite{StHilaire2000} and day of the year \cite{Zhu.2019}. 
The most promising of those parameters is the day of the year because it improves the performance and does not require additional effort in data collection. 
Another parameter, the runoff, was found to usually play a relatively small role in water temperature prediction but shows to be increasingly important for high-altitude catchments \cite{Zhu.2019}.\\
To the best of our knowledge, the current best RMSE values vary between 0.46\textdegree C \cite{Zhu.2019} and 1.58\textdegree C \cite{HadzimaNyarko.2014} in NN approaches over different streams, NN architectures, and input parameters.\\

\vspace{-4mm}
\subsection{Explainable AI}
While all the metrics for evaluating classical regression models, like mentioned above, determine the difference between prediction and observation, NN need more thorough analysis. It is an open problem to understand and interpret their exact behavior, which is why they go by the name of \textit{black-box} models.

Among others, some techniques visualize patterns in input images that trigger the NN \cite{simonyan2014deep}, evaluate their sensitivity to certain inputs \cite{zurada1994sensitivity}, calculate Integrated Gradients \cite{sundararajan2017axiomatic}, or perform layer-wise relevance propagation \cite{bach2015pixel} which is specifically interesting for inputs that are images.
A broad overview of many possible techniques can be found in  \cite{MONTAVON20181}.\\

Almost all of the recent works focus on classification networks whose interpretability is hard to determine.
However, for regression networks, this problem is different.
In our case, there is just one continuous output value that has to be observed and not categorical labels.
That is why we use the gradients of the input values, similar to \cite{sundararajan2017axiomatic}, as a method to visualize the behavior of the NN. 
\subsection{Verification}
NN are naturally very susceptible to adversarial attacks, as many works have demonstrated in recent years \cite{papernot2016limitations, szegedy2014intriguing}.
Consequently, various verification techniques for NN are being developed these days. 
Most of these focus on proving the robustness of the NN \cite{DBLP:conf/atva/Ehlers17, singh2019abstract}. 
Local robustness is usually defined for classification networks:
On a small area around each input to the NN, it should still predict the same label as for the original. 
On regression networks, this property cannot be defined in the same way.
We cannot expect the NN to predict the same value for a slightly perturbed input.
We still want to bind this prediction error in a small region around the inputs.
Unfortunately, many tools for verification only support classification networks.
That is why we decided to use a variant of \textit{DeepPoly} \cite{singh2019abstract}.
This tool uses an abstraction of all possible input values, namely a particular neighborhood, and propagates this through the network.
In the end, it results in an interval of possible output values of the NN.
	\section{Our Approach}
\subsection{Data} \label{data}
We used the following values as possible input features for the NN:
\begin{itemize}
	\item \textit{air temperature [\textdegree C]}:  daily mean values, measured in the distance of 3.25 to 47.03 km from water temperature measurement sites
	\item \textit{runoff [$m^{3}/s$]}: the part of the effective precipitation that flows into the streams and rivers, measured as the amount of water per unit time at each of the selected measurement sites
	\item \textit{day of the year}: the date represented by a value in the interval of 0 to 365
\end{itemize}
The daily mean air temperature [°$C$] 
is provided by the German Meteorological Center (Deutscher Wetterdienst, abbr. DWD) \cite{CDC}. 
The daily mean water temperature [°$C$] and runoff [$m^{3}/s$] are provided by the "Gew\"{a}sserkundlicher Dienst Bayern" (abbr. GkD) \cite{.20210622T10:28:31.000Z}, a department of the Bavarian Environmental Agency. 

The data are free of cost and accessible via download \cite{CDC, .20210622T10:28:31.000Z}. 
For our experiments, we use the data from six German streams selected by the mean annual runoff $MQ \leq \ $1 $m^{3}/s$ and a minimum of 1500 data points. 
For each stream, there is one measurement site for water temperature and runoff by the GkD and one to four surrounding measurement sites for air temperature 
by the DWD. 
In our case, runoff measurements are conducted by the GkD. Hence we can access the data easily.
Still, especially thinking about future predictions, it might not be recommendable to include runoff as it is another value that has to be measured or predicted. 
However, the importance of runoff in water temperature prediction models has already been shown.
It seems to increase the performance for high-altitude catchments \cite{Zhu.2019}, which one should keep in mind while considering it as an input parameter. 
Since not all measurement devices were set up at the same time, measurement periods vary between six and 24 years, depending on the measurement site.

\subsection{Models}
The data structure is quite simple, which is why using complex NN is unnecessary.
In the course of this work, we use 
fully connected NN, which have at most three hidden layers and 90 hidden neurons in total. 
Several previous approaches showed satisfactory performance, even though they used far simpler models than NN. 
That is why we can use smaller models with a modest number of layers and neurons. 

For the training of the NN, we used a particular subset of input features, as described in Section~\ref{data}.

\subsection{Model Analysis}
Regression models, not only NN, are usually evaluated by metrics that compare measured values with predicted values (see Section~\ref{sota_wtp}). In this work, we use the RMSE (as defined in \cite{Rabi15}) as a representative for these metrics because it is a usual and intuitive measure.\\
For simple models, e.g., linear regression or simple stochastic models, the behavior is well-defined and well-known and therefore does not require thorough analysis.
The behavior of NN, on the other hand, is not well-known yet. 
To address this issue, we introduce three additional analyzing methods. These are supposed to help to understand the behavior of the NN. 
\subsubsection{Robustness Analysis}
Local robustness, as defined in \cite{DBLP:conf/cav/KatzBDJK17}, is usually concerned with classification networks and the prediction of labels. Since our approach is based on a regression problem, we do not check whether labels are identical but the following:
Given an input and its neighborhood, how much does the value of the prediction of the NN change at most in this area? 
More formally:
If we have a NN $f:X\rightarrow \mathbb{R}$ that is defined on its input set $X$ and predicts values in $\mathbb{R}$, and an $\epsilon$ defining the size of the neighborhood, we want to get a $\delta$ such that for a subset of inputs of interest $D$:
\begin{equation}
\forall x\in D\subset X\;\;\forall y\in N_\epsilon(x)\;\;:\;\;|f(y)-f(x)|<\delta
\end{equation}
where $N_\epsilon$ defines the $\epsilon$-neighborhood.

If the robustness is low, even a small perturbation to the input values can result in a big difference in the output values.
The tool that we are using to evaluate the robustness of our NN is called \textit{DeepPoly} \cite{singh2019abstract}.
It was originally designed to check the local robustness of classification networks but we adapted it to fit our application. 
\subsubsection{MinMaxAnalysis}
On a simple model, such as a fitted sine function, its minimum and maximum values are obvious and can easily be derived by calculus. 
However, this is not as simple for NN.
For them, it is neither obvious nor an easy calculation.
We adapt the idea from \cite{szegedy2014intriguing} and use gradient descent. 
The principle is similar to backpropagation for learning a NN.
Instead of learning optimal weights to minimize the loss function, we optimize the input vector to minimize resp. maximize the output value, while keeping the values in a reasonable range.
That is, -45\textdegree C to 60\textdegree C for temperature, and -1 to 40 $m^{3}/s$ for the runoff.
This yields input vectors with either very high or very low output values. 
Since the starting position of gradient descent can heavily influence its result, we start the optimization process from a set of randomly chosen input vectors, and we look at the minimum resp. maximum observed output value.

\subsubsection{Impact Analysis}
What we call \textit{impact analysis} is similar to \textit{sensitivity analysis}, which is used to determine which input the NN is sensitive on \cite{zurada1994sensitivity}.
The key idea is to measure how much an input feature contributes to the calculation of the output value.
This can be done by calculating the gradient of the input features based on the loss function. 
This value indicates how much a change in the input feature will affect the computation of the output value.
On the one hand, it gives an idea of which features are necessary, and on the other hand, it helps in inspecting the differences between the streams. 
If the influence of the parameters on the streams is different, this could indicate a relationship between the streams that is unknown so far.
	\section{Experiments}
With the help of a vast grid search, we found that the use of the default hyper-parameters is suitable for accurate water temperature prediction in all cases. 
To be able to use \textit{DeepPoly}, we chose ReLU as an activation function. 
For our fully connected NN, we determined three hidden layers to return consistently good results based on the grid search.
Based on this selection, we compare combinations of several input parameters with different numbers of nodes per layer, using normalized input data.
The dataset is split into test-, train- and validation data; resp. 25\%, 7.5\% and 67.5\% of the whole data.
The optimizer for the training is L-BFGS that is run on a total of 10000 epochs. 

In Section~\ref{sec:featureselection}, we exemplarily fix the architecture to 9-7-13 nodes, since this performs well according to robustness in Section~\ref{sec:robustness}. On the other hand, we fix input parameters in Section~\ref{sec:robustness} to those showing the lowest RMSE values in Section~\ref{sec:featureselection}.

\vspace{-1mm}
\subsection{Input feature selection} \label{sec:featureselection}
\begin{table}[tbp] 
	\caption{RMSE values for NN with different input combinations.\\
		a: air temperature. r: runoff. d: day of the year.}
	\begin{center}
		\begin{tabular}{l | rrrr }
			\toprule
			Stream     & A & A+R    & A+D    & A+R+D  \\ \midrule
			\textbf{One airstation}\\ 
			
			Aubach       & 0.84 & 0.84 & 0.49 & \textbf{0.48}\\
			Abens    & 0.88 & 0.69 & 0.72 & \textbf{0.55} \\
			Bernauer Ache     & 1.02 & 0.91 & 0.83 & \textbf{0.70} \\
			Grosse Ohe   & 1.10 & 0.96 & 0.68 & \textbf{0.59}\\
			Otterbach & 1.74 & 1.72 & 1.57 &  \textbf{1.54} \\
			Sulzbach    & 1.09 & 1.06 & 0.92 &  \textbf{0.91}  \\\midrule
			
			\textbf{All airstations}\\
			
			Aubach       & 0.79 & 0.79 & 0.48 & \textbf{0.47}\\
			Abens    & 0.87 & 0.68 & 0.67 & \textbf{0.52} \\
			Bernauer Ache     & 1.00 & 0.92 & 0.83 & \textbf{0.67} \\
			Grosse Ohe   & 0.97 & 0.51 & 0.80 & \textbf{0.48} \\
			Otterbach & 1.70 & 1.68 & 1.57 &  \textbf{1.57} \\
			Sulzbach    & 1.07 & 1.02 & 0.91 &  \textbf{0.89} \\\bottomrule
			
		\end{tabular}
	\end{center}
	\label{combinations}\vspace{-4mm}
\end{table}
In Table~\ref{combinations} we show a subset of our results, with the combinations of our input values as follows:
only air temperature (A), air temperature and runoff (A+R), air temperature and day of the year (A+D), and the combination of all three of them (A+R+D).
Additionally, we present a comparison between only having one weather station for the prediction, and as many as there are in the direct neighborhood of the stream.\\
We can observe that the obtained RMSE values are already similar to those determined in \cite{Zhu.2019}, if we use only air temperature as input. 
For most streams, it decreases if we add runoff as an input parameter.
The RMSE decreases even more if we combine air temperature and day of the year as an input.
Moreover, the best RMSE values are obtained if the combination of all three input parameters A+R+D is used. 
On the other hand, using all weather stations decreases the RMSE in only three out of six streams for the input combinations A+D and A+R+D. 
For the combination A+R it decreases the RMSE in already five out of six streams and for only air temperature as input, even all six streams show decreased RMSE values. 

In summary, we can observe that the best RMSE values are obtained if all available input parameters are used. 
However, we obtain comparable results with the A+D input combination for several streams, confirming that the runoff does not necessarily have a big impact. 
Moreover, even among the best RMSE values obtained with the A+R+D input combination, we find a high variety in prediction accuracy between the different streams, ranging between 0.47\textdegree C and 1.56\textdegree C. 
We conjecture that there are stream-dependent parameters that we do not use in this work and that should be investigated further by hydrologists. 

\vspace{-1mm}
\subsection{Robustness Analysis}\label{sec:robustness}
While the robustness analysis is meant to be an evaluation method, we already use it to choose a good architecture by training three NN with different architectures for each stream, using the best input parameter combinations as identified in Section~\ref{sec:featureselection}. 
We evaluate the different NN on the RMSE and the NN's robustness. 
To test the robustness, we add a perturbation of 0.01 to the air temperature and discharge which corresponds to approximately 1\textdegree C perturbation on the temperature and 0.4 $m^3/s$ on the discharge. 
The perturbation analysis generates the minimum and maximum possible outcome for the perturbed input, as displayed in Table~\ref{robustness}:
On the one hand, bigger NN usually generate higher possible perturbations. 
On the other hand, for the biggest NN (30-30-30), the RMSE is best in five out of six streams. 
This confirms that choosing NN architecture should not be based on the RMSE alone, since it is inconclusive in terms of NN robustness. 
As we aim to find a good balance between prediction accuracy and perturbation behavior, we use the medium-sized architecture (9-7-13) for further analysis.
\begin{table}[tbp]
	\caption{Different NN architectures (\textbf{shape}) for each \textbf{stream} with corresponding \textbf{RMSE}, average difference after perturbation (\textbf{MeanPerturb}) and Minimum and Maximum values that the best NN can take for each stream.}
	\begin{center}
		\begin{tabular}{l c | r | r | rr}
			\toprule
			Stream                      & Shape    & RMSE & MeanPerturb   & Max   &    Min \\ \midrule
			Aubach                      & 4-5-6    & 0.66 & 0.99          &       &        \\
			Aubach                      & 9-7-13   & 0.47 & \textbf{0.7}  & 49.52 &  -1.96 \\
			\smallskip
			Aubach        				& 30-30-30 & \textbf{0.44} & 0.82          &       &        \\
			Abens                       & 4-5-6    & 0.63 & \textbf{0.78} &        & \\
			Abens                       & 9-7-13   & 0.52 & 1.1           &  43.78 & -19.70        \\
			\smallskip
			Abens        				& 30-30-30 & \textbf{0.50} & 1.7           &       &        \\
			Bernauer Ache               & 4-5-6    & 0.83 & \textbf{1.04} & & \\
			Bernauer Ache               & 9-7-13   & 0.67 & 1.19          &       42.38 & -10.71        \\
			\smallskip
			Bernauer Ache 				& 30-30-30 & \textbf{0.63} & 1.62          &       &        \\
			Grosse Ohe                  & 4-5-6    & 0.53 & 1.18          &       &        \\
			Grosse Ohe                  & 9-7-13   & 0.48 & \textbf{1.07} & 50.14 &  -0.39 \\
			\smallskip
			Grosse Ohe    				& 30-30-30 & \textbf{0.47} & 1.18          &       &        \\
			Otterbach                   & 4-5-6    & 1.70 & 1.45          &       &        \\
			Otterbach                   & 9-7-13   & \textbf{1.54} & \textbf{1.36} & 48.82 &  -5.63 \\
			\smallskip
			Otterbach     				& 30-30-30 & 1.55 & 1.44          &       &        \\
			Sulzbach                    & 4-5-6    & 1.05 & 1.66          &       &        \\
			Sulzbach                    & 9-7-13   & 0.89 & \textbf{1.29} & 61.16 &  -0.36 \\
			Sulzbach     				& 30-30-30 & \textbf{0.81} & 3.12          &       &        \\ \bottomrule
		\end{tabular}
	\end{center}
	\label{robustness}\vspace{-6mm}
\end{table}
\subsection{MinMax-Analysis}
When using unrealistic input values, e.g. temperatures of -45\textdegree C and 60\textdegree C for each of the two weather stations for the last three days combined with very low discharge on January 1$^{st}$, we can obtain very high and low values. For instance, for Sulzbach, this combination leads to a maximum value of 61.16\textdegree C (see Table~\ref{robustness}). 
These input value combinations are artificial and unrealistic. 
Still, we observe that all minimum and maximum values displayed in Table~\ref{robustness} can only be obtained with input values at the boundary of the allowed input values, in our case -45\textdegree C and 60\textdegree C for air temperature, and -1 to 40 $m^{3}/s$ for runoff. 
To apply this approach in the hydrological context for guaranteeing that the NN does not predict unreasonable values, we have to define reasonable values for each relevant waterbody first. 
With this information, the MinMax-Analysis is a powerful tool to test the NN's reliability, especially in the context of prediction of future water temperatures. 

\begin{center}
	\begin{figure}[tbp]
	    \begin{center}
		\includegraphics[width=0.4\textwidth,trim={0 0 0 1cm}]{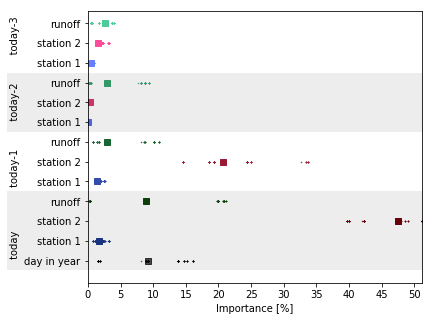}
		\end{center}
		\caption{Impact analysis for Aubach. The x-axis shows the importance for the prediction of each value on the y-axis. The dots represent all values that have occurred in the analysis, the squares mark the median of those. Stations 1 and 2 represent air temperatures, measured at the respective weather stations.}
		\label{impact}
	\end{figure}
\end{center}\vspace{-4mm}
\subsection{Impact Analysis}
For the impact analysis, we have chosen the model of Aubach as an example, since this model performed best in all categories. 
It is depicted in Figure~\ref{impact}. 
On the x-axis, the respective value's importance for the prediction is represented. 
On the y-axis, all input features of the NN are displayed.
The dots represent all of the important values that have occurred, the square indicates their median.
Comparing with the feature selection, we can confirm that besides air temperature the day of the year has an impact on the prediction, as well as the runoff. 
Interestingly, the air temperature values of weather station two have a much higher impact on the result than the values of weather station one, confirming the only low decrease of RMSE between the use of one vs. the use of all stations as seen in Table \ref{combinations}.
	\section{Conclusion and Future Work}
We show that precise water temperature predictions for German streams are possible with NN-based models. 
However, the precision of the resulting NN varies widely, which is an open point for further examination.
Furthermore, the best achieved RMSE of 0.44 is still not as small as the usual measurement inaccuracy, which is at 0.3\textdegree C. 
A future goal is thus to improve the precision to a point where the model is just as good as a direct measurement or at least close enough.
We confirm that the day of the year is an important factor, and so is the runoff.
Additionally, we apply methods already known for classification problems in an adapted version to our regression problem.
This improves the understanding of the behavior of NN for users that are not familiar with it, especially their limitations and their sensitivity to changes.
With this, we contribute to more reliable water temperature predictions that will be applied to different climate change scenarios.\\
Besides the improvement of the training methods, we suggest finding additional easily accessible data on streams that might improve water temperature prediction from a hydrological perspective.
One could also make use of satellite images to include vegetation patterns as an additional input parameter.
Also, we think it would be beneficial to use our methods of assessment, especially the impact analysis, and further examine whether and why certain input features are more important in some models, and less in others.  
Additionally, the heterogeneity between the different streams should be examined further, for which our methods will contribute to an interdisciplinary step forward. 
Concerning the analysis of the NN, the possibilities are not yet exhausted.
There are various other verification tools, that can be used to check certain properties of the NN, e.g. \cite{katzshort}.
Some tools try to evaluate NN-based systems which could be used, e.g. \cite{syrenn}, to have a more precise evaluation of the minimal and maximal value of the NN.
	
 	\section*{Acknowledgment}

This work was financially supported by the DFG Research Training Group on Continuous Verification of Cyber-Physical Systems (GRK 2428) and the AquaKlif project of the Bayklif network funded by the Bavarian State Ministry of Science and Arts (Bayerisches Staatsministerium für Wissenschaft und Kunst).
We also want to thank Jan K\v{r}et\'{i}nsk\'{y}, Romy Wild, and Lisa Kaule for their continuous support. 
	
	\bibliographystyle{IEEEtran}
	\bibliography{bibliography.bib}
	
	\section{Appendix}
\subsection{Dataset}
The dataset was collected from different sources: The daily mean air temperature is provided by the German Meteorological Center (Deutscher Wetterdienst, abbr. DWD) \cite{CDC}. 
The daily mean water temperature and runoff are provided by the "Gew\"{a}sserkundlicher Dienst Bayern" (abbr. GkD) \cite{.20210622T10:28:31.000Z}, a department of the Bavarian Environmental Agency.
\subsubsection{Preprocessing}
As a first step, the data is checked to filter for missing values.
Sometimes, values of -999 are put in the table if there is no measurement available. 
When such a value appears in one column of the dataset, the complete row that contains it is deleted.
Additionally, the values of the DWD and GkD are matched together based on the date to build a complete dataset.
Afterward, the data is normalized to fit a range from 0 to 1. 
\begin{itemize}
    \item The day in the year is a value from 1 to 365 and is thus scaled by $1/365$
    \item The air temperatures in Germany vary between -37.8\textdegree C and 40.2\textdegree C. Due to climate change, the variation may be bigger, so we scaled the values by a minimum of -45\textdegree C and 60\textdegree C
    \item The runoff is generally a value greater than 0, however, to allow for measurement problems, we asume a minimum of $-1$ and a maximum of $40$
\end{itemize}
\subsubsection{Size of the dataset}
\begin{table}[!htb] 
	\caption{The shape of the dataset for the different configurations.\\
	a: air temperature. r: runoff. d: day of the year.}
	\label{datashape}
	\begin{center}
	\begin{tabular}{l|ccccc}
	
	\toprule
    Stream &Number of Datapoints&\multicolumn{4}{c}{Number of Features}\\
    &&A&A+R&A+D&A+R+D\\
    \midrule
    Aubach & 2377 & 8 & 12 & 9 & 13 \\
    Abens & 3566 & 12 & 16 & 13 & 17 \\
    Bernauer Ache & 4607 & 12 & 16 & 13 & 17 \\
    Grosse Ohe & 5080 & 16 & 20 & 17 & 21 \\
    Otterbach & 8534 & 8 & 12 & 9 & 13 \\
    Sulzbach & 4156 & 8 & 12 & 9 & 13 \\
    \bottomrule
\end{tabular}
\end{center}
\end{table}
In Table \ref{datashape}, we point out the sizes of the datasets. 
The number of data points indicates the number of rows in the dataset, after the preprocessing step.
The number of features is indicated for each of the configurations that were used in the experiments: only air temperature (A), air temperature and runoff (A+R), air temperature and day of the year (A+D), and the combination of all three of them (A+R+D).

\subsection{Training}
The networks were created and trained in Python with the package Scikit-learn \cite{pedregosa2011scikit}.
To determine the best working parameters, we did a grid-search on 
\begin{itemize}
    \item the activation function: ReLU, Sigmoid, Tanh
    \item the number of hidden layers: 1 to 3
    \item the number of neurons: 1 to 10, 20, 30, 40, 50, 60, 70, 80, 90
    \item the optimizer: L-BFGS, SGD or ADAM
    \item the learning rate: 0.0001, 0.001, 0.01 (only for SGD and ADAM)
    \item the behavior of the learning rate: constant, scaling, adaptive (only for SGD and ADAM)
\end{itemize}
We just performed the grid search on one dataset.
As activation function, the tanh performed best, however, most verification tools rely on the usage of the ReLU. 
Therefore, and because the difference in the best RMSE is below 0.005, we decided to use the ReLU anyway.\\
The maximum number of iterations was set to 100000. However, the training procedure stops when the validation loss does not improve by at least $10^{-4}$ over 10 epochs.
Despite the fact that second-order optimization procedures, to which L-BFGS can be counted, is usually computationally difficult this is not the case in our application.
This is most likely due to the small size of our networks. 
In the grid search, we found, too, that L-BFGS always provides us with a better loss than SGD or ADAM in all cases. 
Additionally, L-BFGS comes with the advantage that there is no need for a learning rate, so we chose to use it for all the training.

\end{document}